\documentclass[lettersize,journal]{IEEEtran}
\usepackage{amsmath,amsfonts}
\usepackage{algorithmic}
\usepackage{algorithm}
\usepackage{array}
\usepackage[caption=false,font=normalsize,labelfont=sf,textfont=sf]{subfig}
\usepackage{textcomp}
\usepackage{stfloats}
\usepackage{url}
\usepackage{verbatim}
\usepackage{graphicx}
\usepackage{cite}
\usepackage{multirow,makecell,hhline,amsfonts,arydshln}
\usepackage{color}
\usepackage{hyperref}
\hypersetup{colorlinks=true,
linkcolor=blue,
anchorcolor=blue,
citecolor=blue}

\begin{document}

\title{Focus Through Motion: RGB-Event Collaborative Token Sparsification for Efficient Object Detection}
% Focus Through Motion: RGB-Event Collaborative Token Sparsification for Efficient Object Detection
% {Nan~Yang,~Yang~Wang,~Zhanwen~Liu,~Xiangmo~Zhao}
\author{{Nan~Yang$^{\ast}$,~Yang~Wang$^{\ast}$,~Zhanwen~Liu$^{\dagger}$,~Yuchao~Dai,~Yang~Liu,~Xiangmo~Zhao}
\thanks{$^{\ast}$Co-first author. $^{\dagger}$Corresponding author.}
\thanks{Nan Yang, Yang Wang, Zhanwen Liu and Xiangmo Zhao are with the School of Information Engineering, Chang'an University, Shaanxi, Xi’an 710000, China (e-mail: 2022024001@chd.edu.cn; ywang120@chd.edu.cn; zwliu@chd.edu.cn; xmzhao@chd.edu.cn).}
\thanks{Yuchao Dai is with the School of Electronics and Information, Northwestern Polytechnical University and Shaanxi Key Laboratory of Information Acquisition and Processing, Xi’an 710129, China (e-mail: daiyuchao@nwpu.edu.cn).}
\thanks{Yang Liu is with the School of Vehicle and Mobility, Tsinghua University, Beijing 100084, China, and also with State Key Laboratory of Intelligent Green Vehicle and Mobility, Beijing 100084, China (e-mail: thu\_ets\_lab@tsinghua.edu.cn).}
}
        % <-this % stops a space

\markboth{Journal of \LaTeX\ Class Files,~Vol.~14, No.~8, August~2021}%
{Shell \MakeLowercase{\textit{et al.}}: A Sample Article Using IEEEtran.cls for IEEE Journals}


\maketitle

\begin{abstract}
Existing RGB-Event detection methods process the low-information regions of both modalities (background in images and non-event regions in event data) uniformly during feature extraction and fusion, resulting in high computational costs and suboptimal performance. To mitigate the computational redundancy during feature extraction, researchers have respectively proposed token sparsification methods for the image and event modalities. However, these methods employ a fixed number or threshold for token selection, hindering the retention of informative tokens for samples with varying complexity. To achieve a better balance between accuracy and efficiency, we propose FocusMamba, which performs adaptive collaborative sparsification of multimodal features and efficiently integrates complementary information. Specifically, an Event-Guided Multimodal Sparsification (EGMS) strategy is designed to identify and adaptively discard low-information regions within each modality by leveraging scene content changes perceived by the event camera. Based on the sparsification results, a Cross-Modality Focus Fusion (CMFF) module is proposed to effectively capture and integrate complementary features from both modalities. Experiments on the DSEC-Det and PKU-DAVIS-SOD datasets demonstrate that the proposed method achieves superior performance in both accuracy and efficiency compared to existing methods. The code will be available at \url{https://github.com/Zizzzzzzz/FocusMamba}.
\end{abstract}


\section{Introduction}

\IEEEPARstart{O}{bject} detection is fundamental to intelligent systems like autonomous driving and intelligent security \cite{falanga2020dynamic,gehrig2024low,liu2024mstf,liu2023multi,li2024intention,li2023regional,peng2024lightweight,liu2021cascade,zhao2021end,liu2024multi,kuang2024harnessing,qu2023envisioning,li2022two}. In poor lighting (\emph{e.g.}, low-light and over-exposure) and fast motion scenes, the low dynamic range and frame rate of frame-based cameras cause low-quality images\cite{peng2025boosting,peng2025pixel}, significantly degrading the performance of frame-based methods \cite{liu2024enhancing,liu2024boosting,wang2017deep,wang2023decoupling,peng2024efficient,wang2023brightness,wu2025robustgs,wang2020deep}. Recently, the advent of event cameras has provided a promising solution to this challenge. With the high temporal resolution and high dynamic range, the event camera exhibits stable imaging under challenging conditions \cite{gallego2020event,ding2023mlb,peng2024unveiling,tan2022multi,wan2024event,chen2022progressivemotionseg,wang2019progressive,xie2024event,chen2022ecsnet,liu2023voxel}. However, they perform poorly in static or slow-motion scenes and cannot capture color and texture, which are critical for object detection \cite{li2023sodformer,long2024spike,liu2025beyond,liu2024event}. Therefore, researchers \cite{tomy2022fusing,li2023sodformer,liu2024enhancing,yang2023joint,cao2024chasing} have proposed various fusion frameworks that combine the strengths of both cameras to improve robustness in challenging conditions.

\begin{figure}[tbp]
\centering
\includegraphics[scale=0.385]{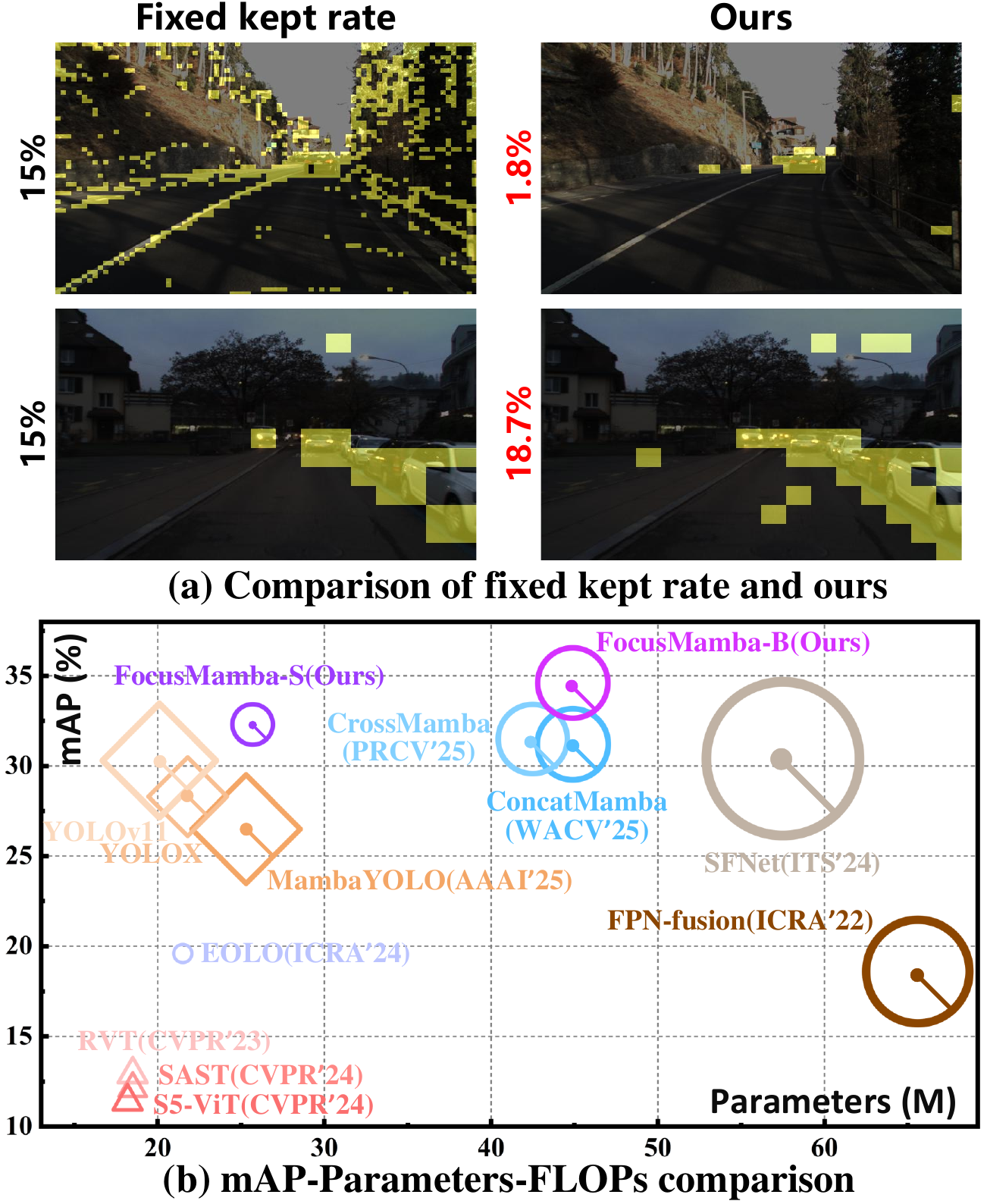}
\caption{(a) Comparison of token retention (yellow regions) between the fixed kept rate and our method. Our method can adaptively retain an appropriate number of tokens based on the scene complexity. (b) The mAP-Parameters-FLOPs comparison between state-of-the-art methods and our FocusMamba on the DSEC-Det dataset. The triangles, diamonds, and circles represent event-based, frame-based, and fusion-based methods, respectively, with the radius of each shape corresponding to its FLOPs. Our FocusMamba achieves an excellent balance between accuracy and efficiency.}
\label{fig1}
\end{figure}

Existing fusion frameworks independently extract representations for two modalities, followed by fusion modules to capture complementary information and improve representation discriminability. Although these methods outperform single-modal methods in accuracy, they inevitably reduce efficiency for two primary reasons. Firstly, during the representation extraction stage, they treat all information of both modalities equally, including crucial object areas and low-information regions (\emph{e.g.}, backgrounds in images and no-event regions in event data), neglecting the high redundancy of image backgrounds \cite{roh2021sparse,liu2022adaptive,liu2024revisiting} and the spatial sparsity of event data \cite{peng2024scene,yang2025smamba}. This leads to significant redundant computation and high computational costs. To address this issue, \cite{roh2021sparse,chen2023sparsevit,liu2024revisiting} and \cite{peng2024scene,yang2025smamba} have respectively proposed token sparsification methods for the image and event modalities, discarding low-information regions from computation. However, their token selection strategy based on a fixed number or threshold neglects variations in sample content, which may result in under-sparsification of simple samples or over-sparsification of complex ones. As shown in Fig.\ref{fig1}(a), at a 15\% token retention rate, existing methods preserve excessive background content in the simple scene while failing to retain critical object information in the complex scene. Secondly, during the fusion stage, they apply attention mechanisms across all regions of both modality features to capture complementary relationships. However, this introduces background interference, leading to inaccurate perception of complementary regions and redundant computations.

% 当采用15%的token保留率时，现有方法会在简单的场景保留过多的背景，在复杂的场景丢失重要的目标信息

To address the above issues, this paper proposes an RGB-Event collaborative sparsification object detection framework, FocusMamba, which aims to leverage the complementary characteristics of both modalities to perform adaptive sparsification of multimodal features and efficient integration of complementary features, achieving an excellent balance between accuracy and efficiency, as shown in Fig.\ref{fig1}(b). First, we design an Event-Guided Multi-modality Sparsification (EGMS) strategy that adaptively discards low-information regions for each sample based on scene content changes perceived by the event camera, significantly reducing computational load, as shown in Fig.\ref{fig1}(a). Specifically, a modality-specific statistical scoring method independently evaluates the importance of each modality's tokens, maximizing the strengths of each modality. Subsequently, the event spatial ratio, which reflects the information content of objects, is utilized to adjust the score differentiation and threshold for each sample, generating sparsification maps that guide subsequent selective scanning and MLP for sparsification calculations. The sparsification maps effectively reflect the informative regions within each modality. The differences between these regions, arising from the distinct perceptual strengths of the modalities, highlight inter-modal complementary relationships. Building upon this, we propose a Cross-Modality Focus Fusion (CMFF) module that first efficiently captures complementary regions by perceiving the differences between the sparsification maps of the two modalities, and then performs fine-grained integration in the important areas preserved by EGMS, thereby suppressing background interference and reducing redundant computation. Overall, our contribution can be summarized as follows:

(1) We propose the FocusMamba, which adaptively discards low-information regions for each sample, and focuses on the modeling and fusion of informative regions, achieving an ideal balance between accuracy and efficiency.

(2) We design the EGMS strategy, which effectively identifies informative regions across modalities and adaptively retains them for each sample, significantly reducing computational costs.

(3) We propose the CMFF module, which effectively perceives and finely integrates complementary features across modalities based on sparsification results, thereby eliminating background interference.

(4) Comparison with 12 state-of-the-art methods (including event-only, RGB-only, and RGB-Event fusion) on the DSEC-Det and PKU-DAVIS-SOD datasets demonstrates that our method achieves the best performance, surpassing the second-best method SFNet by 4.2\% and 0.8\%, while requiring only 28.5\% and 22.2\% of its FLOPs, respectively.

% Experimental results on the DSEC-Det and PKU-DAVIS-SOD datasets demonstrate that our method surpasses state-of-the-art methods, achieving superior performance.

% 在DSEC-Det and PKU-DAVIS-SOD数据集上，与12种SOTA方法，包括事件单模态、RGB单模态以及多模态方法，的对比结果表明，我们的方法实现了最好的性能，分别超越次优方法SFNet4.2%和0.8%，仅需其28.5%和22.2%的FLOPs.

\section{Related Work}
% This section provides a comprehensive overview of event-based and RGB-Event fusion-based object detection methods, followed by a review of recent advancements in token sparsification methods and Vision Mamba.
This section provides an overview of RGB-Event fusion based detection methods, followed by a review of recent advancements in token sparsification and Vision Mamba.
% \textbf{Event-based object detection.} RED and ASTMNet integrate CNNs and RNNs to extract spatiotemporal representations from event streams. RVT introduces the self-attention mechanism to improve spatiotemporal modeling. SSM-ViT proposes a state-space model (SSM) with learnable time-scale parameters to fully exploit the high temporal resolution of event cameras, enabling high inference frequencies. SAST leverages the spatial sparsity of event data and proposes an adaptive sparsification strategy to reduce computational costs. However, event cameras are limited in capturing stationary objects, color, and fine-grained texture information which are essential for high-performance object detection. As a result, the performance of these methods remains inferior to that of frame-based approaches.
% Event-based detection methods.事件相机的优良特性受到了研究人员的广泛关注，大量工作将其应用于目标检测任务。RED和ASTMNet提出聚合CNNs和RNNs以从事件流中提取时空表征。RVT引入了自注意力机制并进行精细化的模块设计以更高效地建模时空表征，实现了精度和效率之间的平衡。SSM-ViT提出具有可学习时间尺度参数的状态空间模型(SSM)以充分利用事件相机的高时间分辨率特性，使得模型能够适应不同的推理频率。SAST利用事件数据的空间稀疏特性，提出自适应稀疏化策略来减少无事件区域的计算冗余。最近，FARSE-CNN提出Fully Asynchronous, Recurrent and Sparse CNN以利用事件数据的异步特性，显著提升了计算效率，但其性能远低于基于CNN和Transformer的方法。尽管研究人员已经付出了大量的努力来发掘事件相机在目标检测任务中的潜力，但由于事件相机不能捕获静止目标，并且缺乏颜色、细致纹理这些对高性能目标检测至关重要的信息，这些方法的性能依旧低于基于帧的方法。
% The unique advantages of event cameras have attracted significant attention from researchers, leading to numerous applications in object detection tasks. 

\textbf{RGB-Event fusion for object detection.} Fusing RGB and event modalities is a crucial strategy for achieving high-performance detection in challenging conditions\cite{tomy2022fusing,zhou2022rgb,li2023sodformer,liu2024enhancing,cao2025embracing}. Among these fusion methods, the feature-level fusion strategy effectively utilizes the complementary strengths of both modalities, advancing rapidly. FPN-fusion \cite{tomy2022fusing} fuses features through concatenation. However, simple concatenation fails to differentiate the degradation levels of each modality and balance their contributions adaptively. To address this, SODFormer \cite{li2023sodformer} perceives the degraded features of each modality through self-attention mechanisms to achieve adaptive fusion. SFNet \cite{liu2024enhancing} completes information lost in images by cross-modality global spatial similarity calculation. CAFR \cite{cao2025embracing} achieves coarse-to-fine feature fusion through the cross-attention mechanism. However, they process the low-information regions of both modalities uniformly, resulting in high computational redundancy and inaccurate capture of complementary features. 

% \textbf{Token sparsification.}
% Recently, token sparsification methods have been proposed to improve the computational efficiency of the representation modeling process. In token scoring, they assign token importance scores using L2 norm of features \cite{chen2023sparsevit,liu2022adaptive} or by designing learnable modules for token importance prediction \cite{rao2021dynamicvit,kong2022spvit,peng2024scene}. In token selection, they retain important tokens by a fixed number \cite{chen2023sparsevit,rao2021dynamicvit} or threshold \cite{liu2022adaptive,peng2024scene}, reducing the computational cost of the self-attention mechanism. However, the token selection strategy based on the fixed number or threshold has poor adaptability to diverse scenarios, which may result in under-sparsification of simple samples or over-sparsification of complex ones.

\textbf{Token sparsification.}
Recently, token sparsification methods have been proposed to improve the computational efficiency of the VLM (Vision-Language Model) domain, image classification, and object detection. It is noted that most sparsification methods in the VLM domain \cite{xing2024pyramiddrop,chen2024image,ye2025fit,dhouib2025pact,lin2025boosting} rely on the ViT framework, text modality, or prior knowledge specific to VLM tasks, making them unsuitable for dense detection task and the Mamba framework. In image classification and object detection, they assign token importance scores using statistical methods (L2 norm \cite{chen2023sparsevit,liu2022adaptive} and spatial-temporal continuity of events \cite{yang2025smamba}) or by designing learnable modules for token importance prediction \cite{rao2021dynamicvit,kong2022spvit,liu2024revisiting,peng2024scene} in token scoring stage. In token selection, they retain tokens by a fixed number \cite{chen2023sparsevit,rao2021dynamicvit,liu2024revisiting} or threshold \cite{liu2022adaptive,peng2024scene}, reducing the computational cost. However, the token selection strategy based on the fixed number or threshold has poor adaptability to diverse scenarios, hindering the retention of informative tokens for samples with varying complexity.

\textbf{Vision Mamba.} Compared to Transformers, Mamba \cite{mamba} achieves a better balance between global modeling capability and computational efficiency. Specifically, Mamba introduces the input-dependent selective scanning mechanism to improve global modeling, while proposing a parallel scanning method to maintain linear complexity. Building on this, ViM \cite{zhu2024vision} and VMmamba \cite{liu2024vmamba} have developed the core Vision Mamba module to explore Mamba's potential in visual tasks. Researchers \cite{xu2024survey,patro2024mamba,wan2024sigma,guo2024mambair,zou2024wave,huang2024mamba,peng2025directing,di2025qmambabsr} have subsequently applied it to various visual tasks, demonstrating its advantages in the vision domain. However, Mamba lacks effective mechanisms for handling redundant image backgrounds and sparse event data, resulting in substantial redundant computations.

\section{Method}

\begin{figure*}[tbp]
\centering
\includegraphics[scale=0.79]{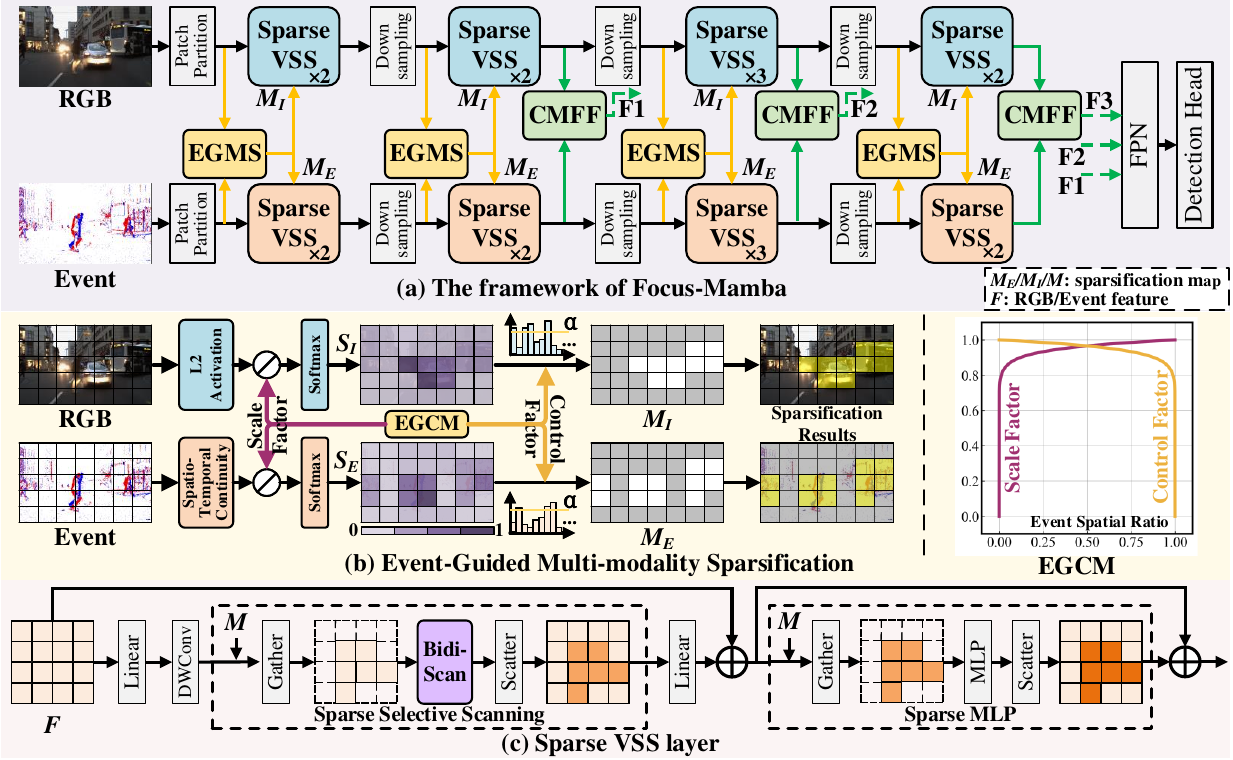}
\caption{\textbf{The architecture of FocusMamba.} The image and event tensors are first tokenized and then processed through four stages for multi-scale feature extraction. Before each stage, EGMS generates independent sparsification maps for both modalities to guide the sparsification operation. The CMFF module is applied in the final three stages to fuse the features of two modalities.}
\label{fig2}
\end{figure*}

\subsection{The Overview of the FocusMamba}
The framework of FocusMamba is illustrated in Fig.\ref{fig2}(a). Initially, the event stream is converted into voxel tensor \cite{zhu2019unsupervised}, and both the voxel tensor and RGB image are divided into patches for tokenization. Subsequently, the tokens from the two modalities are independently fed into two separate backbones, which process them through four stages to extract multi-scale features. In each stage, the Event-Guided Multi-modality Sparsification (EGMS) strategy, as shown in Fig.\ref{fig2}(b), adaptively generates independent sparsification maps ($M$) for both modalities. These sparsification maps guide sparse selective scanning and sparse MLP in the subsequent Sparse VSS layer, significantly reducing computational costs, as shown in Fig.\ref{fig2}(c). The Cross-Modality Focus Fusion (CMFF) module, as shown in Fig.\ref{fig4}, is placed after the last three stages to effectively capture and integrate complementary features from both modalities under the guidance of the sparsification maps. The fused features are then fed into the FPN for multi-scale feature fusion. Finally, the YOLOv8 detection head outputs the detection results.
% ASFMamba的网络架构如图1所示。首先，我们follow先前的工作[1]，将事件流转换为事件帧，并将事件帧和RGB图像划分为patch块进行token化。随后，两个模态的token分别输入到两个独立的骨干网络中，并通过4个stages处理来生成多尺度特征。在每个stage中，Event-Guided Multi-modality Sparsification (EGMS)模块，如图2所示，在事件流的引导下自适应地为两个模态生成独立的稀疏化map。这些稀疏化map M在后续的VSS layer中引导稀疏选择性扫描和稀疏MLP操作，显著减少了计算量，如图1（b）所示。Cross-Modality Feature Enhancement (CMFE)模块被放置在后三个stage来对两个模态的互补特征进行有效捕捉和整合。融合后的特征被输入到FPN中进行多尺度特征融合，最后由Yolov8检测头输出检测结果。

\subsection{Event-Guided Multi-modality Sparsification}
To adaptively discard low-information regions, we propose an Event-Guided Multi-modality Sparsification (EGMS) strategy, which consists of the scoring module and Event-Guided Control Mechanism (EGCM), as shown in Fig.\ref{fig2}(b).
% 为了自适应地从计算中去除两个模态的低信息区域，我们提出一种Event-Guided Multi-modality Sparsification (EGMS)策略，如图2所示。在token评分方面，可学习的token评分模块会由于早期和中间层token的表征编码不足，难以识别出重要的token。因此，我们采用更可靠的统计评分方法来独立地评价每个模态token的重要性，以有效识别各模态的重要信息，同时维持了最小的计算代价。在token选择方面，我们设计了Event-Guided Control Mechanism (EGCM)，它基于事件相机对场景内容变化的感知能力自适应地选择两个模态保留的token的数量。实现了样本级的多模态自适应稀疏化，极大地降低了计算量。

% \begin{figure*}[tbp]
% \centering
% \includegraphics[scale=0.83]{EGMS-0103.pdf}
% \caption{\textbf{The architecture of EGMS strategy.} To facilitate observation, a large patch size is used for tokenization. The scores of tokens from both modalities are independently assessed using statistical scoring methods. EGMS then adjusts the contrast of token scores and selection thresholds based on the event spatial ratio, thereby adaptively retaining important information (yellow regions) for both modalities.}
% \label{fig3}
% \end{figure*}

% 两个模态token的重要性通过统计评分方法来独立衡量，EGMS随后基于事件空间占比对不同token得分的对比度和token选择的阈值进行调控，从而自适应地为两个模态保留下重要的信息。

%我们从稀疏和密集场景中分别选取10个token来展示它们在scale前后的分值差异。scale factor能够有效放大token间的分值差异，尤其是在稀疏场景中。

% Guided by the sparsification maps, it accurately captures and fully utilizes the complementary features of both modalities, while eliminating background interference.
% 在两个模态稀疏化map的引导下，CMFF module对两个模态的互补特征进行精确捕捉和充分利用，消除了背景的干扰并减少了冗余计算。

\subsubsection{Scoring Module}
Learnable scoring modules often fail to identify important tokens due to insufficient representation encoding in the early and middle layers \cite{chefer2021transformer,wu2020visual}. Therefore, we adopt more reliable statistical scoring methods \cite{chen2023sparsevit,yang2025smamba} to independently assess the importance of tokens in each modality, effectively identifying key information for each modality while maintaining minimal computational cost.

% 对于图像模态，采用L2 activation作为评估指标，它能够有效衡量每一个token的特征响应幅值和重要性，从而识别出目标区域

% 对于图像模态，L2 activation能够有效衡量每一个token的特征响应幅值和重要性，从而识别出目标区域，因此采用L2 activation作为评估指标

% For the image modality, the L2 activation of each token is employed as its importance score. L2 activation has been proven as a more effective metric for evaluating feature importance in both CNNs \cite{he2018soft} and ViTs \cite{liu2022adaptive,chen2023sparsevit}.

% For the image modality, the L2 activation is employed as the scoring metric. L2 activation effectively measures the feature response magnitude and importance of each token, thereby identifying the object regions.

% For the image modality, 目标区域期望被保留，背景区域期望被去除。因此，采用L2 activation作为评分指标，它能够effectively measures the feature response magnitude and importance of each token \cite{he2018soft,liu2022adaptive,chen2023sparsevit}, enabling the identification of object regions.
% For the image modality, L2 activation is employed as the scoring metric, as it effectively measures the feature response magnitude and importance of each token \cite{he2018soft,liu2022adaptive,chen2023sparsevit}, enabling the identification of object regions.

For the image modality, object regions are expected to be retained, while background regions are to be discarded. Therefore, L2 activation is employed as the scoring metric, as it effectively measures the feature response magnitude and importance of each token \cite{he2018soft,liu2022adaptive,chen2023sparsevit}, enabling the identification of object regions. Specifically, the L2 norm $S_I\in\mathbb{R}^N$ of image tokens $X\in\mathbb{R}^{N\times C}$ is computed as:
\begin{equation} 
{S_I}_i=\|X_i\|_2=\sqrt{\sum_{j=1}^CX_{i,j}^2},
\end{equation}
where $N$ is the token number, $C$ is the feature dimension.

% For the event modality, 活动事件区域期望被保留，噪声和无事件区域期望被去除。因此，the spatiotemporal continuity of events is used as the scoring metric. 它能够utilizing the differences in spatiotemporal distributions between active and noisy events，有效识别出活动事件区域
% For the event modality, the spatiotemporal continuity of each region is used as its importance score. This metric effectively distinguishes important event tokens by utilizing the differences in spatiotemporal distributions between active and noisy events.

For the event modality, activity event regions are expected to be retained, while noise and non-event regions are to be discarded. Therefore, the spatiotemporal continuity of events \cite{yang2025smamba} is used as the scoring metric, as it effectively identifies activity event regions based on the spatiotemporal distribution differences between activity and noisy events. Specifically, given the event stream $\left\{\left(x_{i}, y_{i}, t_{i}, p_{i}\right)\right\}_{i=1}^{K}$, where $(x_i,y_i)$ is the spatial coordinates, $t_i$ is the timestamp, and $p_i\in\{{-1,1}\}$ is the polarity. First, to identify temporally discontinuous noise events, the timestamps of events within the region corresponding to each token are accumulated to assess the temporal continuity, as follows:
\begin{equation}
{S_E^{T}}_{x,y}=\sum_{i,x_{i}=x,y_{i}=y}t_{i}, \\
S_E^{T}=Maxpooling(S_E^{T}),
\end{equation}
the kernel size and stride of MaxPooling are set to $P$, where $P$ is the downsampling rate of the current stage. Subsequently, to further identify spatially isolated noise events and maintain complete structures for objects with sparse events, a Gaussian kernel is used to aggregate neighborhood information and assess the spatiotemporal continuity score, as follows:
\begin{equation}
S_E=\frac{\sum_{q\in\Omega}\left(\exp\left(-\frac{\parallel q-c\parallel^2}{2\sigma^2}\right){S_E^{T}}_q\right)}{\sum_{q\in\Omega}\exp\left(-\frac{\parallel q-c\parallel^2}{2\sigma^2}\right)},
\end{equation}
where $c$ is the center of the neighborhood $\Omega$, ${S_E^{T}}_q$ is the value of the neighborhood $q$, and $\sigma$ represents the variance. 

\begin{figure}
\centering
\includegraphics[scale=0.31]{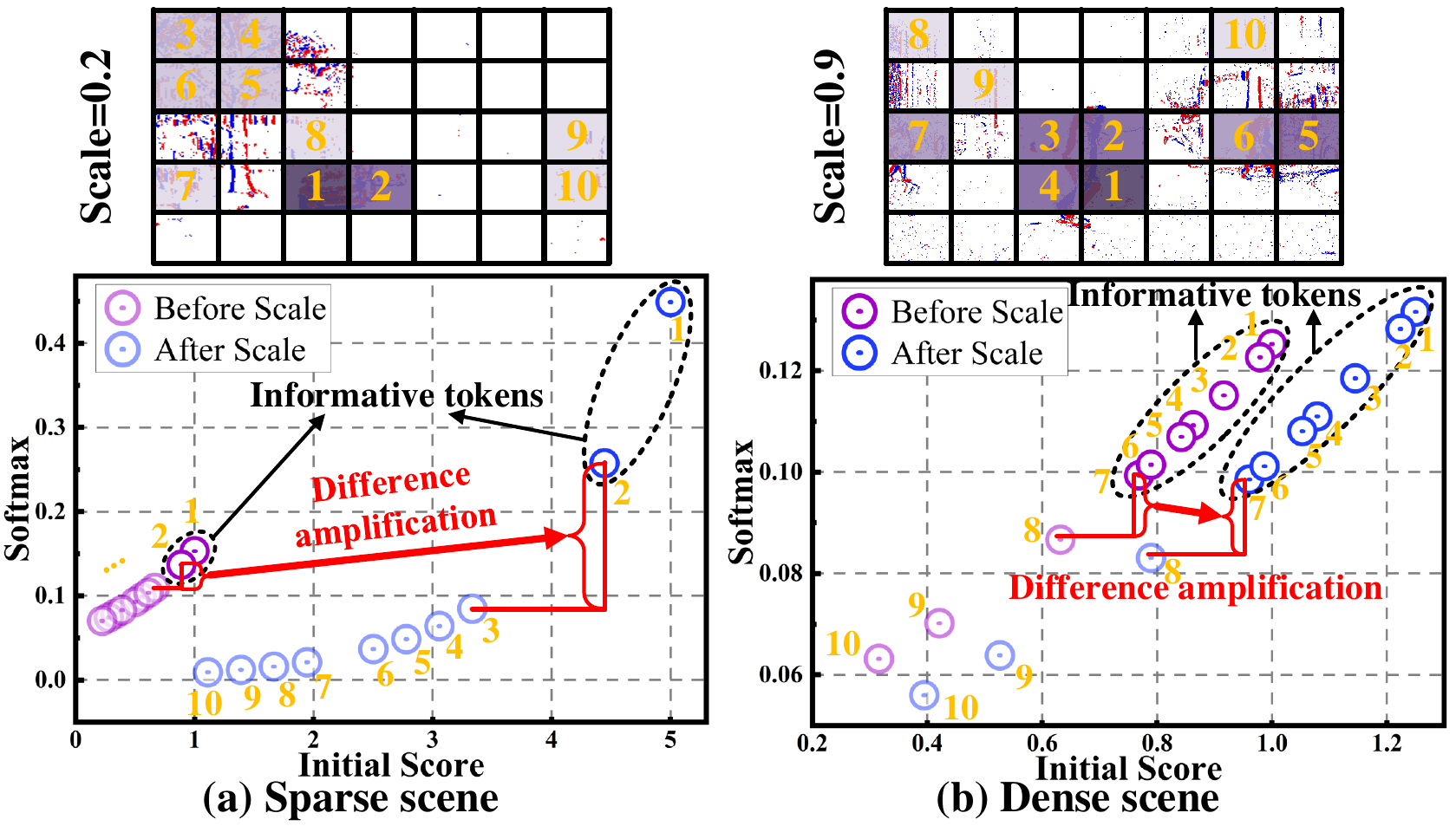}
\caption{We select 10 tokens from sparse and dense object-information scenes to show the score differences before and after scaling. The scale factor effectively amplifies the score differences, particularly in scenes with sparse object information.}
\label{fig3}
\end{figure}

\subsubsection{Event-Guided Control Mechanism}
% 在获得token的评分图$S_R$和$S_E$，一种直接的方法是通过设置specific的token保留数量或者分数阈值来实现稀疏化。但这会面临两个问题:(1)The initial scores derived from the scoring module show insufficient contrast for effective differentiation高信息token和低信息token; (2)固定的数量或阈值无法适应场景内容的变化，为不同场景保留下重要token
After obtaining the token score maps $S_I$ and $S_E$, a straightforward method for sparsification is to set a specific retention number or score threshold. However, this method has two limitations: (1) the initial scores from the scoring module lack sufficient contrast to distinguish informative tokens from low-information ones effectively; and (2) fixed retention numbers or thresholds fail to adapt to scene content variations, making it difficult to consistently preserve informative tokens across diverse scenarios. To address these challenges, we propose the Event-Guided Control Mechanism (EGCM), as shown in Fig.\ref{fig2}(b), which adaptively regulates the score differences and selection thresholds based on the scene's sparsification level perceived by event camera, achieving sample-level multi-modal adaptive sparsification.

Moving objects frequently cause brightness changes along edges, continuously triggering events. Consequently, we can estimate the spatial proportion of object information by computing the ratio $r$ of pixels that trigger events to the total number of pixels. The ratio $r$ can reflect the objects' information content and the scene's sparsification level to some extent. Based on this, we design the \textbf{scale factor} to amplify inter-token differences and the \textbf{control factor} to adaptively regulate the selection threshold.
% However, directly computing the event spatial ratio $r$ is susceptible to event noise, leading to inaccurate estimation of object information. To address this, we count pixels that trigger events only within regions whose scores exceed the average value of $S_E$.
% 事件相机通过捕捉亮度变化来检测场景内容的变化，the event spatial ratio $r$ which quantifies the ratio of pixel with brightness changes 能够在一定程度上反映目标

% 事件相机能够通过捕捉目标运动产生的亮度变化来感知场景中目标的信息含量。

% In sparser scenes, the object and potential background regions are more likely to be grouped together in tokens with high scores. To effectively distinguish the object from the background, a greater contrast between these tokens is necessary. 
% In sparse scenes, the score difference between important and low-information tokens is small, which hinders effective distinction between them. Therefore

\textbf{Scale factor.} The scale factor is integrated with softmax to amplify the score differences among tokens, as shown in Fig.\ref{fig3}. The process is as follows:
\begin{equation}
Scale=r^{\frac{1}{\rho}},
\end{equation}
\begin{equation}
S_I=softmax(\frac{S_I}{Scale}),S_E=softmax(\frac{S_E}{Scale}),
\end{equation}
$\rho$ is a hyperparameter that regulates the sparsification process. In scenes with sparser object information, the scale value decreases, thereby amplifying token scores and increasing the disparity between them after the softmax operation. This facilitates the identification of informative tokens and the filtering of low-information ones.

% To effectively select important tokens To adaptively retain important tokens based on the complexity of the scene, we also design a control factor based on $r$. This factor adjusts the sparsification threshold $\alpha$ according to the scene's sparsity level.
\textbf{Control factor.} The control factor adaptively adjusts the sparsification threshold $\alpha$ according to the scene's sparsity level. The process is as follows:
\begin{equation}
Control=(1-r)^{\frac{1}{\rho}}, \alpha=\frac{1}{N}\times Control.
\end{equation}
% \begin{equation}
% \alpha=\frac{1}{N}\times Control.
% \end{equation}

% In sparse scenes, the number of tokens containing valid information is limited, resulting in higher scores after softmax compared to dense scenes
In scenes with sparse object information, the limited number of informative tokens leads to higher softmax scores compared to dense scenes, as shown in the vertical axis of Fig.\ref{fig3}. As a result, a higher threshold is required to filter out potential background regions. In contrast, scenes with dense object information contain more informative tokens, which results in lower softmax scores. Therefore, a lower threshold is necessary to prevent the loss of important information. Sparsification maps for both modalities are generated based on the threshold $\alpha$:
\begin{equation}
M_I=\left\{\begin{array}{c}1,\mathrm{if}\ S_I\geq\alpha,\\0,\mathrm{if }\ S_I<\alpha,\end{array}\right.M_E=\left\{\begin{array}{c}1,\mathrm{if}\ S_E\geq\alpha,\\0,\mathrm{if}\ S_E<\alpha.\end{array}\right.
\end{equation}

The sparsification maps guide the sparsification operations in the subsequent layers of this stage.

\begin{figure*}[tbp]
\centering
\includegraphics[scale=0.95]{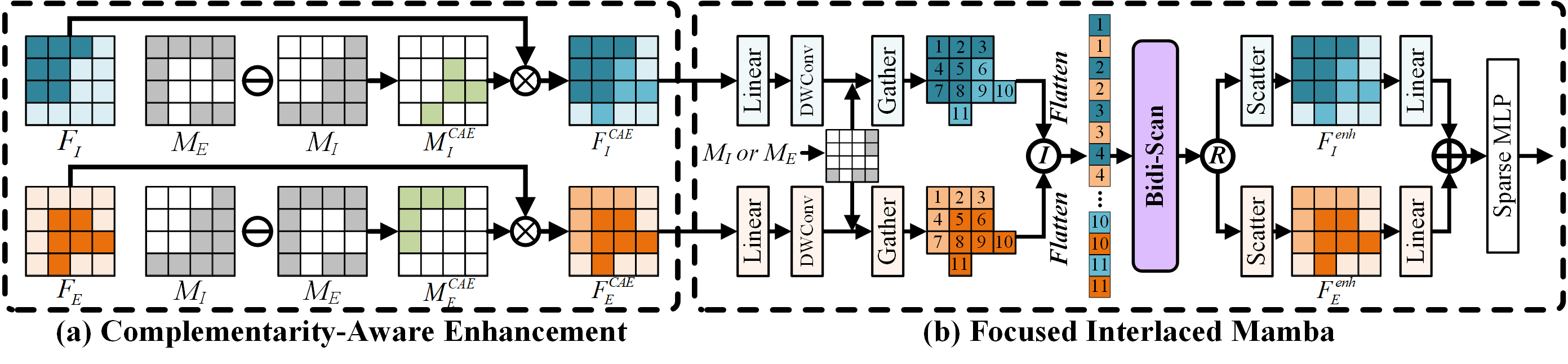}
\caption{\textbf{Cross-Modality Focus Fusion module.} Guided by the sparsification maps $M_I$ and $M_E$, the CMFF module accurately captures and fully leverages the complementary features, eliminating background interference and reducing redundant computation.}
\label{fig4}
\end{figure*}

\subsection{Cross-Modality Focus Fusion}

The regions retained in the sparsification maps primarily consist of high-quality information captured by the two modalities based on their respective perceptual strengths. For example, frame-based cameras can capture rich textures and static objects, while event cameras are capable of detecting objects under challenging exposure conditions. Due to the distinct perceptual advantages of the two modalities, their retained regions in the sparsification maps $M_I$ and $M_E$ are different, as shown in Fig.\ref{fig5}(a). These differences highlight the inter-modal complementary relationship. Building on this, we propose the Cross-Modality Focus Fusion module, consisting of a Complementarity-Aware Enhancement (CAE) module and a Focused Interlaced Mamba (FI-Mamba) module, as shown in Fig.\ref{fig4}. Guided by the sparsification maps, it accurately captures and fully utilizes the complementary features of both modalities, while eliminating background interference.

To enhance the degraded regions in each modality (\emph{e.g.}, improperly exposed regions in RGB images and static object regions in event data) caused by perceptual limitations, we design the CAE module, as shown in Fig.\ref{fig4}(a), which efficiently captures and utilizes inter-modal complementary regions. Specifically, the degradation regions of one modality relative to the other are perceived by performing the logical operation on the sparsification maps $M_I$ and $M_E$, which then guides the enhancement of its features. Taking the image modality as an example, the process is as follows:
% \begin{equation}
% F_{R}^{CAE}=\beta\cdot(M_{E}\ XOR\ (M_{E}\ AND\ M_{R}))\cdot F_{R},
% \end{equation}
% \begin{equation}
% F_{E}^{CAE}=\beta\cdot(M_{R}-M_{E})\cdot F_{E},
% \end{equation}
\begin{equation}
M_{E}-M_{I}=M_{E}\ xor\ (M_{E}\ and\ M_{I}),
\end{equation}
\begin{equation}
M_{I}^{CAE}=\left\{\begin{array}{c}\beta,\mathrm{if}\ M_{E}-M_{I}=1,\\1,\mathrm{if }\ M_{E}-M_{I}=0,\end{array}\right.
\end{equation}
% \begin{equation}
% M_{I}-M_{E}=M_{I}\ xor\ (M_{E}\ and\ M_{I}),
% \end{equation}
\begin{equation}
F_{I}^{CAE}=M_{I}^{CAE}\cdot F_{I}, 
\end{equation}
% F_{I},F_{E}^{CAE}=\beta\cdot(M_{I}-M_{E})\cdot F_{E},
% \begin{equation}
% F_{I}^{CAE}=\beta\cdot(M_{E}-M_{I})\cdot F_{I},F_{E}^{CAE}=\beta\cdot(M_{I}-M_{E})\cdot F_{E},
% \end{equation}
% \begin{equation}
% M_{E}-M_{R}=M_{E}\ {\land}\ (M_{E}\ \&\ M_{R})
% \end{equation}
where $xor$ is the logical exclusive OR operation, $and$ is the logical AND operation, $\cdot$ denotes element-wise multiplication. $M_{I}^{CAE}$ and $\beta$ are the enhancement map and coefficient, respectively.
% $\beta$ is an enhancement coefficient.
% $-$ is the set difference operation,

% \subsubsection{Focused Interlaced Mamba}

% Specifically, the features from both modalities are preprocessed using Linear and DWConv. Then, the important tokens are selected based on the union operation of the sparsification maps $M_R$ and $M_E$, eliminating background interference while preventing the loss of crucial information:
% \begin{equation}
% % M=M_{E}\cup M_{R},
% M=M_{E}\ or\  M_{R},
% \end{equation}
% \begin{equation}
% F_{R}^{\prime}=gather\left(F_{R}^{CAE},M\right),F_{E}^{\prime}=gather\left(F_{E}^{CAE},M\right),
% \end{equation}
% where $or$ is the logical or operation, the gather operation selects $F_{R}^{\prime}$ and $F_{E}^{\prime}$ from $F_{R}^{CAE}$ and $F_{E}^{CAE}$ based on the indices of non-zero elements in $M$. 

To further fully leverage the complementary advantages of both modalities, we propose the FI-Mamba, as shown in Fig.\ref{fig4}(b), which integrates important tokens and simultaneously models long-range dependencies within and across modalities, facilitating refined feature fusion. Specifically, the features from two modalities are preprocessed using Linear and DWConv. Then, the important tokens are indicated based on the logical OR operation of the sparsification maps $M_E$ and $M_I$:
\begin{equation}
% M=M_{E}\cup M_{R},
M=M_{I}\ or\  M_{E},
\end{equation}
% \begin{equation}
% F_{R}^{\prime}=gather\left(F_{R}^{CAE},M\right),F_{E}^{\prime}=gather\left(F_{E}^{CAE},M\right),
% \end{equation}
where $or$ is the logical OR operation. After obtaining $M$, the gather operation selects important tokens $F_{I}^{\prime}$ and $F_{E}^{\prime}$ from $F_{I}^{CAE}$ and $F_{E}^{CAE}$ based on the indices of non-zero elements in $M$, eliminating background interference while shortening the scan distance between crucial information. Subsequently, the interleaving operation flattens the selected tokens and alternately merges them into a new sequence. The sequence is input into the Bidi-Scan mechanism \cite{liu2024vmamba} to simultaneously model global dependencies both within and between modalities, enabling feature enhancement by leveraging information from both intra-modal and inter-modal sources. After that, the scatter operation maps the enhanced features $F_{I}^{enh}$ and $F_{E}^{enh}$ back to the original feature map based on the indices of non-zero elements in $M$. Finally, the features are added and input into the Sparse MLP to further refine.

\section{Experiments}
% This section begins with an overview of the experimental setup. Subsequently, a comparative analysis of our method against state-of-the-art (SOTA) methods is presented. The visualization results are then shown to demonstrate the scene-adaptability of our method. Finally, ablation studies are conducted to validate the effectiveness of our approach. 
This section begins with an overview of the experimental setup, followed by a comparative analysis of our method against state-of-the-art (SOTA) approaches. Then, ablation studies are conducted to validate the effectiveness of our method. Finally, we present visualization results to demonstrate the scene adaptability of our method.

\subsection{Experimental Setup}\label{Experimental Setup}
% This subsection details the employed datasets, the validation metrics utilized, and the implementation details.
\textbf{Datasets.} We conduct experiments on two datasets DSEC-Det \cite{liu2024enhancing} and PKU-DAVIS-SOD \cite{li2023sodformer}. The DSEC-Det dataset contains challenging variable lighting conditions and high-quality manually annotated labels, providing more than 208k bounding boxes for 8 classes. There exist other versions of annotations for this dataset \cite{tomy2022fusing,gehrig2024low}, but they use automatic annotation methods, which result in poor label quality. Therefore, this work utilizes the more comprehensive and accurate annotations provided by \cite{liu2024enhancing}. The PKU-DAVIS-SOD dataset contains motion blur, low-light, and static scenes which is challenging for both modalities, and provides manual bounding boxes at 25 Hz for 3 classes, yielding more than 1080.1k labels. Considering that the data quality of the two modalities in the DSEC-Det dataset is higher and closer to real-world scenarios, this work chooses it to perform ablation studies.

\textbf{Metrics.} The COCO mAP \cite{lin2014microsoft} is used to evaluate the accuracy of the model. To assess model efficiency, we also report parameters, FLOPs, and runtime.

\textbf{Implementation Details.} To ensure a fair comparison, we follow prior methods\cite{liu2024enhancing,tomy2022fusing,li2023sodformer} and sample the event stream using the inter‑frame intervals of the RGB modality (50 ms for DSEC‑Det and 40 ms for PKU‑DAVIS‑SOD). The image size is set to 640×640 for the DSEC-Det dataset and 416×416 for the PKU-DAVIS-SOD dataset during training and testing. We implemented the framework with Pytorch on Ubuntu 20.04 and trained it with 2 × NVIDIA-3090 and a batch size of 5 on each GPU. We adopted the Adam optimizer with the weight decay of 0.0005 and the learning rate of 0.01.

\subsection{Quantitative Results}
% To evaluate the superiority of our method, we compare our FocusMamba with recent SOTA (state-of-the-art) object detection methods on the DSEC-Det and PKU-DAVIS-SOD datasets. Furthermore, to demonstrate the effectiveness of our EGMS strategy, we conduct comparative experiments with other token sparsification methods.
We first compare our FocusMamba with recent SOTA object detection methods on DSEC-Det and PKU-DAVIS-SOD datasets, and then conduct comparative experiments with other sparsification methods to demonstrate the effectiveness of our EGMS strategy.

\begin{table*}[tbp]
\small
\centering
\renewcommand\arraystretch{1.1} 
\caption{Performance comparison on DSEC-Det and PKU-DAVIS-SOD datasets. Values in brackets (·) indicate the percentage decrease in FLOPs compared to the baseline method. The best and the suboptimal performances are marked in \textbf{bold} and \underline{underline}, respectively.}
\scalebox{1.0}{
\begin{tabular}{c|c|c|ccc|ccc|c}
     
    \Xhline{1.5pt}
    {\multirow{2}*{{\textbf{}}}} & {\multirow{2}*{{\textbf{Methods}}}} & {\multirow{2}*{{\textbf{Pub.\&Yea.}}}} & \multicolumn{3}{c}{{\textbf{DSEC-Det}}} & \multicolumn{3}{c}{{\textbf{PKU-DAVIS-SOD}}} & \\
    % \hline
    \hhline{~~~-------}
    % \hline
     & & &{\scriptsize mAP50/mAP} & FLOPs & Runtime & {\scriptsize mAP50/mAP} & FLOPs & Runtime & Params\\
    \hline
    % \hline
    {\multirow{3}*{\rotatebox{90}{Event}}}
    % & ASTMNet & CNN+RNN & - & - & - & -/29.1 & - & 21.3 & \textgreater100M\\
    % & RVT & {CVPR'23} & 25.1/12.9 & 19.6G & 11.9ms & 50.3/25.6 & 6.5G & 7.1ms & 18.5M\\
    & SAST & {CVPR'24} & 24.3/12.1 & 18.5G & 18.8ms & 48.7/24.5  & 6.2G & 16.7ms & 18.5M\\
    & S5-ViT & {CVPR'24} & 22.3/11.4 & 19.5G & 12.3ms & 46.6/23.3 & 7.3G & 10.7ms & 18.2M\\
    & SMamba & {AAAI'25} & 29.0/14.8 & 17.0G & 26.1ms & 53.0/27.1 & 5.2G & 19.0ms & 17.0M\\
    % \hline
    \hline
    {\multirow{3}*{\rotatebox{90}{RGB}}} & YOLOX & arXiv'21 & 43.5/26.5 & 73.8G & 15.3ms & 57.5/30.6 & 31.1G & 10.7ms & 25.3M\\
    & YOLOv11 & arXiv'24 & 46.7/30.3 & 68.0G & 16.2ms & 58.0/30.9 & 44.2G & 15.6ms & 20.1M\\
    & MambaYOLO & AAAI'25 & 44.1/28.3 & 47.2G & 22.2ms & 57.1/29.8 & 21.6G & 21.0ms & 21.8M\\
    \hline
    % \hline
    {\multirow{8}*{\rotatebox{90}{RGB-Event}}} & FPN-fusion & ICRA'22 & 34.4/18.6 & 89.6G & 30.8ms & 36.6/19.5 & 49.7G & 24.0ms & 65.6M\\
    % & RENet & CNN & 37.3/22.2 & - & 17.4ms & 54.9/28.8 & - & 14.2ms & 59.8M\\
    & SODFormer & { TPAMI'23} & - & - & - & 50.4/20.7 & 62.5G & 39.7ms & 82.0M\\
    & EOLO & ICRA'24 & 33.9/19.6 & 13.7G & 330.2ms & 47.2/22.0 & 8.9G & 326.4ms & 21.5M\\
    & SFNet & TITS'24 & 51.4/30.4 & 209G & 44.8ms & 59.6/31.9 & 135.9G & 42.3ms & 57.5M\\
    & CAFR & ECCV'24 & 23.4/12.5 & 213.3G & 117.7ms & 52.0/26.0 & 63.5G & 102.3ms & 82.4M\\
    & ACGR & CVPR'25 & - & - & - & 51.9/- & -G & 16.8ms & 41.4M\\
    % & HDIFormer &{\footnotesize Transformer} & -& -&- & 51.6/- & - & 47.9ms & 64.2M\\
    % & FAOD & -&- &- & 57.5/30.5 & - & 13.6 & 20.3\\
    \hhline{~---------}
    & SFNet* & TITS'24 & 25.0/12.7 & 54.2G & 54.1ms & 46.7/23.0 & 27.4G & 50.9ms & 38.3M\\
    & CAFR* & ECCV'24 & 50.2/30.1 & 52.8G & 50.5ms & 58.0/30.8 & 27.1G & 49.7ms & 39.1M\\
    & ConcatMamba & WACV'25 & 52.2/31.2 & 61.4G & 55.3ms & 58.3/31.1 & 30.4G & 53.3ms & 44.9M\\
    & CrossMamba & PRCV'25 & \underline{52.5}/31.5 & 59.8G & 53.4ms & 58.3/31.2 & 29.3G & 52.7ms & 42.5M\\
    % \rowcolor{gray!20!white}
    & baseline-B & -  & \underline{52.5}/\underline{32.6} & 87.2G & 61.0ms & \underline{59.9}/\underline{32.2} & 42.9G & 57.7ms & 44.9M\\
    % & {\footnotesize FocusMamba-B} & - & \textbf{55.3}/\textbf{34.6} & \thead{60.8G\\{\footnotesize (-30.3\%)}} & 54.7ms & \textbf{60.6}/\textbf{32.7} & \thead{30.2G\\{\footnotesize (-29.6\%)}} & 53.5ms & 44.9M\\
    & {\footnotesize \textbf{FocusMamba-B}} & - & \textbf{55.3}/\textbf{34.6} & 60.8G{ (-30.3\%)} & 54.7ms & \textbf{60.6}/\textbf{32.7} & 30.2G{ (-29.6\%)} & 53.5ms & 44.9M\\
    & {\footnotesize \textbf{FocusMamba-S}} & - & 51.6/32.3 & 35.9G & 53.4ms & 59.7/\underline{32.2} & 17.8G & 52.5ms & 25.7M\\
    \Xhline{1.5pt}
\end{tabular}}
\label{table1}
\end{table*}

\textbf{Comparison with SOTA object detection methods.} We compare our method with recent SOTA methods, including three event-based methods: SAST \cite{peng2024scene}, S5-ViT \cite{zubic2024state} and SMamba \cite{yang2025smamba}; three frame-based methods: YOLOX \cite{ge2021yolox}, YOLOv11 \cite{khanam2024yolov11} and MambaYOLO \cite{wang2024mamba}; as well as six fusion-based methods: FPN-fusion \cite{tomy2022fusing}, SODFormer \cite{li2023sodformer}, EOLO \cite{cao2024chasing}, SFNet \cite{liu2024enhancing}, CAFR \cite{cao2025embracing} and ACGR \cite{li2025asynchronous}. Furthermore, to compare with recent Mamba-based fusion methods and existing RGB-Event fusion modules, we introduce novel baselines by replacing our CMFF module with ConcatMamba \cite{wan2024sigma}, CrossMamba \cite{huang2024mamba}, and the fusion modules of SFNet and CAFR (referred to as SFNet* and CAFR*, respectively). And a baseline model without the EGMS strategy is established to evaluate the effectiveness of the proposed method. For a fair comparison with other methods, we train two models, a base model (FocusMamba-B) and a small model (FocusMamba-S) by adapting the channel dimensions in each stage.
% , achieving the lowest FLOPs and parameter count among fusion-based approaches

As shown in Tab.\ref{table1}, our FocusMamba-B outperforms all other methods. For instance, compared to the SOTA fusion-based method SFNet on the two datasets, our method improves mAP by 4.2\% and 0.8\%, while requiring only 29.1\% and 22.2\% of SFNet's FLOPs, respectively. When compared to our constructed baseline, CrossMamba, our method achieves mAP improvements of 3.1\% and 1.5\%. Furthermore, after incorporating our sparsification strategy into baseline-B, FocusMamba-B achieves a 30.3\% and 29.6\% reduction in FLOPs and a 10.3\% and 7.3\% decrease in runtime on the two datasets, respectively. This strategy also enables the model to focus on more critical regions, thereby improving detection accuracy. When compared to frame-based methods, our small model, FocusMamba-S, achieves higher accuracy and the lowest FLOPs while maintaining similar parameters. To be noted, FocusMamba's inference speed is currently limited by the early-stage development and inefficient hardware utilization of Mamba. Comparative experiments on the two datasets reveal that our method strikes an excellent balance between accuracy and efficiency through multimodal adaptive sparsification and effective integration of complementary features.

\begin{table}[t]
    \centering
        \centering
        \small
        \caption{Performance compared with SOTA token sparsification methods on DSEC-Det dataset.}
        \begin{tabular}{cc|ccc}
        \Xhline{1.5pt}
        {\textbf{Methods}} & {\textbf{Pub.\&Yea.}} & { mAP50/mAP} & FLOPs & Runtime\\ % & Params\\
        \hline
        % \hline
        DivPrune & CVPR'25 & 51.1/30.2 & 68.0G & 60.0ms\\
        % {\footnotesize DynamicViT}  & 52.7/32.6 & 69.5G & 60.4ms\\ % & 46.0M\\
        AS-ViT & IJCAI'23 & 53.7/33.0 & 65.3G & 58.3ms\\
        SparseViT & CVPR'23 & 52.8/32.9 & 69.7G & 55.5ms\\ %  & 44.9M\\
        SViT & ECCV'22 & 53.8/32.8 & 68.5G & 59.0ms\\ %  & 45.6M\\
        SAST & CVPR'24 & 53.2/32.6 & 63.2G & 58.5ms\\ %  & 45.6M\\
        \textbf{Ours} & - & \textbf{55.3}/\textbf{34.6} & 60.8G & 54.7ms\\ %  & 44.9M\\
        \Xhline{1.5pt}
    \end{tabular}
    \label{tab:qua_EGMS}
\end{table}

\begin{figure*}[tbp]
\centering
\includegraphics[scale=0.57]{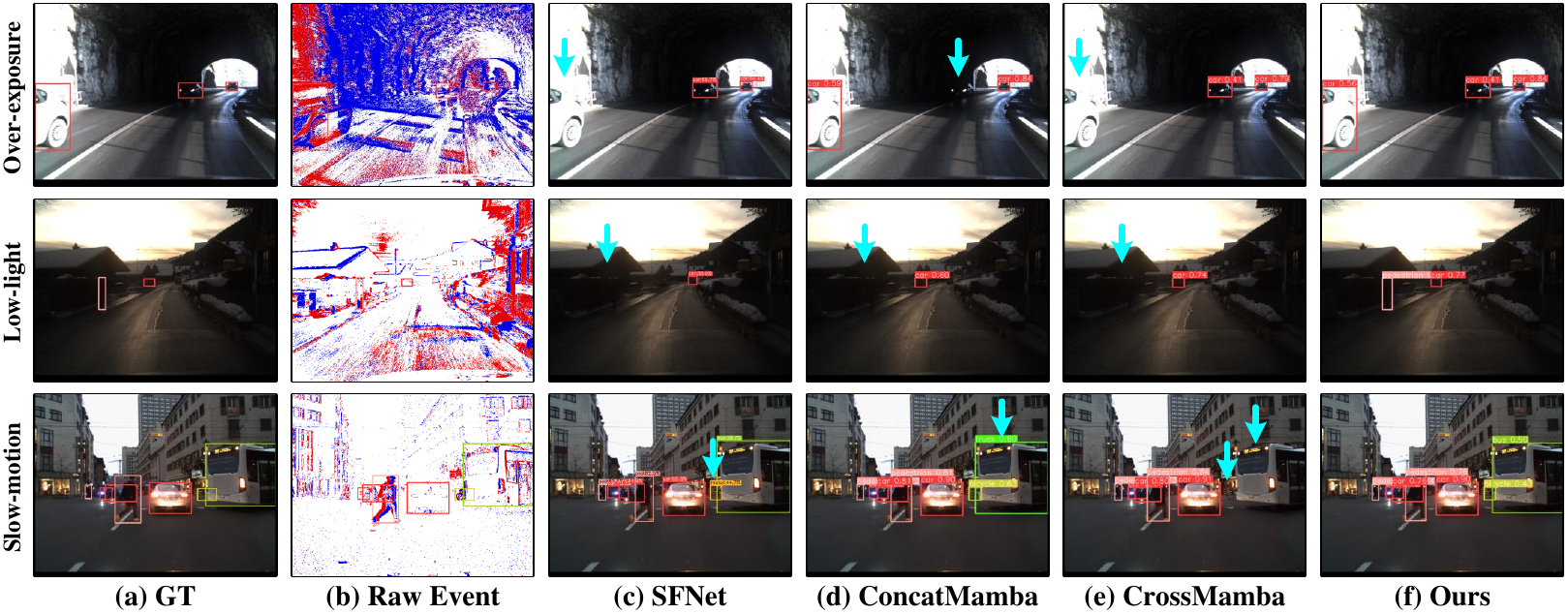}
\caption{Qualitative comparison with SFNet, ConcatMamba, and CrossMamba on the DSEC-Det dataset. We utilize blue arrows to mark the failed cases.}
\label{fig10}
\end{figure*}

\begin{figure*}[tbp]
\centering
\includegraphics[scale=0.57]{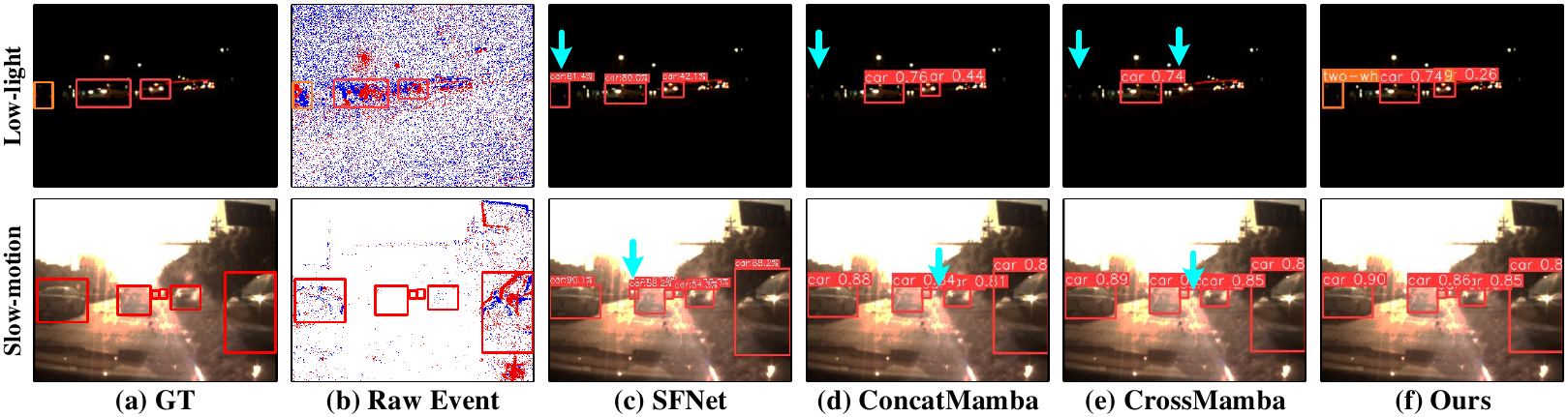}
\caption{Qualitative comparison with SFNet, ConcatMamba, and CrossMamba on the PKU-DAVIS-SOD dataset. We utilize blue arrows to mark the failed cases.}
\label{fig11}
\end{figure*}

\textbf{Comparison with other token sparsification methods.} To further validate the effectiveness of our EGMS, we compare it with one available VLM domain method, DivPrune \cite{alvar2025divprune}; three image-modality methods: AS-ViT \cite{liu2022adaptive}, SparseViT \cite{chen2023sparsevit}, and SViT \cite{liu2024revisiting}; and one event-modality method, SAST \cite{peng2024scene} on the DSEC-Det dataset. For fairness, all methods are adapted to sparsify both modalities independently while maintaining consistent architectural design.
% To align with our method, SAST is modified to adjust scores for both modalities based on event sparsity.
% To further validate the effectiveness of our EGMS, we compare with three image-modality methods (DynamicViT\cite{rao2021dynamicvit}, AS-ViT\cite{liu2022adaptive}, and SparseViT\cite{chen2023sparsevit}) and one event-modality method (SAST\cite{peng2024scene}) on the DSEC-Det dataset. For fairness, all methods are adapted to sparsify both modalities independently while maintaining consistent architectural design. To align with our method, SAST is modified to adjust scores for both modalities based on event sparsity.

As shown in Tab.\ref{tab:qua_EGMS}, the VLM-domain method DivPrune aims to retain the most diverse set of tokens rather than the object-relevant tokens required for detection tasks, resulting in the lowest performance. Despite utilizing L2 activation to improve scoring reliability, AS-ViT and SparseViT underperform due to their fixed threshold and pruning rate. SAST exhibits a degree of scene adaptability, achieving the lowest FLOPs among the compared methods. However, its adaptability is constrained by the unreliable learning-based scoring method and fixed threshold, resulting in limited accuracy. In contrast, our method leverages a reliable scoring mechanism specifically designed to align with the characteristics of modality signals, combined with an effective adaptive regulation strategy, achieving superior adaptive capability and detection accuracy.

\begin{table}[t]
        \centering
        \small
        \caption{Performance of our components.}
        \begin{tabular}{c|ccc}
     
            \Xhline{1.5pt}
            \textbf{Method} & { mAP50/mAP} & FLOPs& Params\\
            % \hline
            % \hhline{~~-----}
            % \hline
             % & & mAP50 & mAP & FLOPs & Runtime & Params\\
            % \hline
            \hline
             & 50.3/31.6 & 81.5G & 37.6M\\
            \hdashline
            + EGMS & 51.2/32.4 & 54.5G & 37.6M\\
            + CAE & 53.6/33.6 & 54.2G & 37.6M\\
            % w/o CAE & 53.4/33.4 & 64.0 & 44.9\\
            + FI-Mamba & \textbf{55.3}/\textbf{34.6} & 60.8G & 44.9M\\
            % with EGMS & 51.2/32.4 & 54.5 & 37.6\\
            % w/o FI-Mamba & 53.6/33.6 & 54.2 & 37.6\\
            % w/o CAE & 53.4/33.4 & 64.0 & 44.9\\
            % with all & \textbf{55.3}/\textbf{34.6} & 60.8 & 44.9\\
            % w/ EGMS & 51.2/32.4 & 54.5 & 37.6\\
            % \thead{w/ EGMS \\and FI-Mamba} & 53.6/33.6 & 54.2 & 37.6\\
            % w/ EGMS and CAE & 53.4/33.4 & 64.0 & 44.9\\
            % w/ all & \textbf{55.3}/\textbf{34.6} & 60.8 & 44.9\\
            % + EGMS & 51.2/32.4 & 54.5 & 37.6\\
            % % \xrowht[()]{00.1pt}
            % \thead{+ EGMS \\ + FI-Mamba} & 53.6/33.6 & 54.2 & 37.6\\
            % \thead{+ EGMS \\ + CAE} & 53.4/33.4 & 64.0 & 44.9\\
            % with all & \textbf{55.3}/\textbf{34.6} & 60.8 & 44.9\\
            % with EGMS & 51.2/32.4 & 54.5 & 37.6\\
            % \thead{with EGMS \\ and FI-Mamba} & 53.6/33.6 & 54.2 & 37.6\\
            % \thead{with EGMS \\ and CAE} & 53.4/33.4 & 64.0 & 44.9\\
            % with all & \textbf{55.3}/\textbf{34.6} & 60.8 & 44.9\\
            \Xhline{1.5pt}
        \end{tabular}
        \label{tab:ablate_component}
\end{table}

% \begin{figure}[tbp]
\begin{figure*}[tbp]
\centering
\includegraphics[scale=0.6]{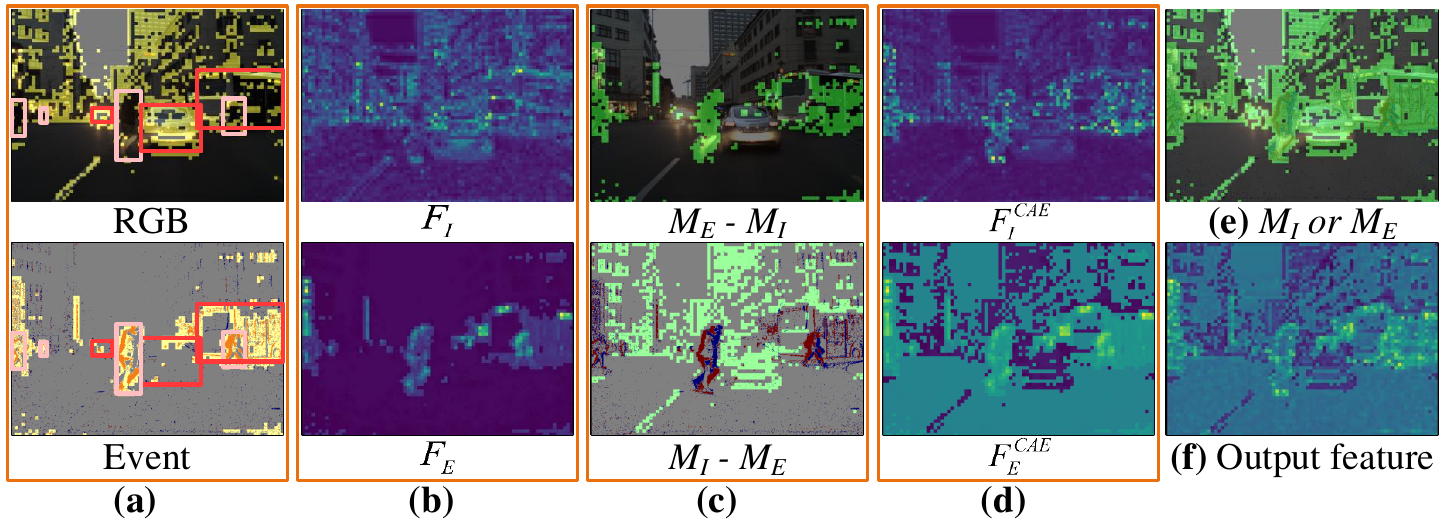}
\caption{Visualization of features before and after CMFF module. (a) illustrates the kept regions of the image and event modalities. (c) represents the difference between the kept regions of the two modalities. (b) and (d) present the features before and after the CAE. (e) shows the union of the kept regions from the two modalities. (f) represents the feature after the FI-Mamba.}
\label{fig5}
\end{figure*}

\subsection{Qualitative Comparison} 
We compare our FocusMamba with SFNet \cite{liu2024enhancing}, ConcatMamba \cite{wan2024sigma}, and CrossMamba \cite{huang2024mamba} on the DSEC-Det and PKU-DAVIS-SOD datasets, as shown in Fig.\ref{fig10} and \ref{fig11}. Our FocusMamba demonstrates more robust detection performance in challenging scenarios (\emph{e.g.}, over-exposure, low-light, and slow-motion), successfully detecting all objects.

\subsection{Ablation Studies}
To assess the effectiveness of the proposed method, we conduct extensive ablation studies based on our base model, FocusMamba-B on the DSEC-Det dataset.
% For further details, please refer to Appendix \ref{More Ablation Studies}.

% \begin{figure*}[htbp]
% \centering
% \includegraphics[scale=0.23]{fea-0106.pdf}
% \caption{Visualization of features before and after CMFF module. (b,c) and (d,i) represent the features before and after the CAE. (j) represent the features after the FI-Mamba.}
% \label{fig5}
% \end{figure*}

\textbf{Contribution of our modules.} We take VSS \cite{liu2024vmamba} as the base backbone to evaluate the contributions of EGMS, CAE, and FI-Mamba modules. As shown in Tab.\ref{tab:ablate_component}, introducing the three modules progressively provides incremental improvements in both performance and efficiency. Specifically, incorporating the EGMS strategy reduces computation and interference from low-information regions, leading to a 33.1\% decrease in FLOPs and a 0.8\% increase in mAP. The addition of the CAE module enhances degraded regions in both modalities, increasing mAP by 1.2\% without additional computational costs. Finally, integrating FI-Mamba further leverages complementary features across modalities, resulting in another 1.0\% improvement in mAP with only a minor computational overhead.

The feature maps before and after the CMFF module are illustrated in Fig.\ref{fig5}. CAE module effectively perceives the complementary regions between modalities, as indicated by the green regions in Fig.\ref{fig5}(c), guiding the image modality to enhance regions with improper exposure and the event modality to suppress low-quality sparse events caused by slow motion, as shown in Fig.\ref{fig5}(d). FI-Mamba effectively captures the complementary dependencies between important regions of the two modalities, as shown in Fig.\ref{fig5}(e), mitigating background interference and modeling more distinguishable object representations, as shown in Fig.\ref{fig5}(f).

% \begin{table}[tbp]
% \small
% \centering
% \caption{Performance of different scoring methods.}
% % \renewcommand\arraystretch{1.2} 
% \begin{tabular}{c|cccc}
     
%     \Xhline{1.5pt}
%     {\textbf{Methods}} & mAP50/mAP & FLOPs  & Params\\
%     % \hline
%     % \hhline{~~-----}
%     % \hline
%      % & & mAP50 & mAP & FLOPs & Runtime & Params\\
%     \hline
%     \hline
%     STCA* & 52.0/32.2 & 61.5 & 44.9\\
%     L2 Norm* & 50.0/31.6 & 55.4 & 44.9\\
%     L2 Norm & 52.7/32.3 & 53.6 & 44.9 & \\
%     SAST & 48.9/30.6 & 51.3 & 45.6\\
%     Ours & \textbf{55.3}/\textbf{34.6} & 60.8 & 44.9\\
%     \Xhline{1.5pt}
% \end{tabular}

% \label{table4}
% \end{table}

\textbf{Scoring module.} To evaluate the effectiveness of our scoring module, we compare it with two joint scoring methods: (1) STCA*: The event modality is scored by the STCA module proposed in \cite{yang2025smamba}, and the images use the same score; (2) L2 Norm*: The sum of the two modalities' features is scored based on L2 norm, and the scoring results are shared; and two independent scoring methods: (1) L2 Norm: Scoring each modality independently based on L2 norm; (2) SAST: Scoring each modality independently based on the learning-based scoring module proposed in \cite{peng2024scene}. As shown in Tab.\ref{tab:ablate_scoring}, compared L2 Norm with L2 Norm*, independent scoring outperforms joint scoring under the same metric. This is because joint scoring fails to retain important information for each modality. However, the inaccuracy of the learning-based scoring module in SAST leads to even lower performance than the joint scoring method. Our method, which employs the modality-specific independent scoring strategy, maximizes the strengths of both modalities and achieves superior performance.

\begin{table}[t]
    \centering
        \centering
        \small
        \caption{Performance of different scoring methods.}
        \begin{tabular}{c|cccc}
             
            \Xhline{1.5pt}
            {\textbf{Methods}} & mAP50/mAP & FLOPs  & Params\\
            % \hline
            % \hhline{~~-----}
            % \hline
             % & & mAP50 & mAP & FLOPs & Runtime & Params\\
            \hline
            % \hline
            STCA* & 52.0/32.2 & 61.5 & 44.9\\
            L2 Norm* & 50.0/31.6 & 55.4 & 44.9\\
            L2 Norm & 52.7/32.3 & 53.6 & 44.9 & \\
            SAST & 48.9/30.6 & 51.3 & 45.6\\
            \textbf{Ours} & \textbf{55.3}/\textbf{34.6} & 60.8 & 44.9\\
            \Xhline{1.5pt}
        \end{tabular}
    \label{tab:ablate_scoring}
\end{table}
\begin{table}[t]
        \centering
        \small
        \caption{Performance of the factors in EGMS, w/o is without.}
        \begin{tabular}{c|cc}
        \Xhline{1.5pt}
        {\textbf{Methods}} & mAP50/mAP & FLOPs\\
        % \hline
        % \hhline{~~-----}
        % \hline
         % & & mAP50 & mAP & FLOPs & Runtime & Params\\
        % \hline
        \hline
        fixed kept rate & 53.0/32.8 & 60.8\\
        w/o scale\&control & 51.2/31.5 & 58.6\\
        with scale & 53.6/33.3 & 58.3\\
        with control & 52.6/32.6 & 61.0\\
        \textbf{with all} & \textbf{55.3}/\textbf{34.6} & 60.8\\
        \Xhline{1.5pt}
    \end{tabular}
    \label{tab:ablate_EGMS}

\end{table}

% \begin{table}[tbp]
% % \small
% \centering
% \renewcommand\arraystretch{1.0} 
% \caption{Performance of the factors in EGMS, w/o is without.}
% \begin{tabular}{c|cc}
%     \Xhline{1.5pt}
%     {\textbf{Methods}} & mAP50/mAP & FLOPs\\
%     % \hline
%     % \hhline{~~-----}
%     % \hline
%      % & & mAP50 & mAP & FLOPs & Runtime & Params\\
%     \hline
%     \hline
%     w/o scale \& control & 49.9/31.5 & 58.6\\
%     with scale & 53.6/33.3 & 58.3\\
%     with control & 52.6/32.6 & 61.0\\
%     Ours & \textbf{55.3}/\textbf{34.6} & 60.8\\
%     \Xhline{1.5pt}
% \end{tabular}
% \label{table5}
% \end{table}

% \begin{wrapfigure}{rt}{0.5\textwidth}
% \centering
% \includegraphics[scale=0.181]{token_vis0508.pdf}
% \caption{Visualizations of sparsification results at different stages.}
% \label{fig6}
% \end{wrapfigure}

\textbf{The factors in EGCM.} We conduct an ablation study to analyze the contribution of the scale factor and control factor in the EGCM. Furthermore, to compare with the fixed kept rate method, we first calculate the average token kept ratios of EGMS and then apply these ratios to remove the same number of tokens for all samples. As shown in Tab.\ref{tab:ablate_EGMS}, without scale and control factors to regulate the sparsification process, the model lacks scene-adaptive capability, resulting in poor performance. The scale factor facilitates the recognition of informative regions and the filtering of low-information areas, improving accuracy and reducing FLOPs. The control factor directly adjusts the sparsification threshold, enabling the model to adaptively select the number of tokens to retain based on the scene's sparsity, surpassing the fixed kept rate method. Finally, combining both factors achieves a significant performance improvement, providing superior adaptability. 

\begin{table}[t]
    \centering

        \centering
        \small
        \caption{Performance of different focus strategies in FI-Mamba.}
        \begin{tabular}{c|cc}
            \Xhline{1.5pt}
            {\textbf{Methods}} & mAP50/mAP & FLOPs\\
            % \hline
            % \hhline{~~-----}
            % \hline
             % & & mAP50 & mAP & FLOPs & Runtime & Params\\
            \hline
            \hline
            Inter-Mamba & 51.3/31.8 & 63.8\\
            IFI-Mamba & 49.0/30.6 & 59.0\\
            \textbf{Ours} & \textbf{55.3}/\textbf{34.6} & 60.8\\
            \Xhline{1.5pt}
        \end{tabular}
        \label{tab:ablate_FIMamba}
\end{table}

\begin{table}[t]
        \centering
        \small
        \caption{Performance of extending EGMS to the ViT-based framework.}
            \begin{tabular}{c|ccc}
                \Xhline{1.5pt}
                {\textbf{Methods}} & { mAP50/mAP} & FLOPs & Runtime\\ % & Params\\
                % \hline
                \hline
                MaxViT  & 46.7/29.8 & 134.7G & 60.5ms\\ %  & 45.6M\\
                \textbf{MaxViT+EGMS} & \textbf{47.4}/\textbf{30.3} & \textbf{73.4}G & \textbf{45.8}ms\\ %  & 44.9M\\
                \Xhline{1.5pt}
            \end{tabular}
            \label{vit}
\end{table}

\begin{figure}[t]
\centering
\includegraphics[scale=0.22]{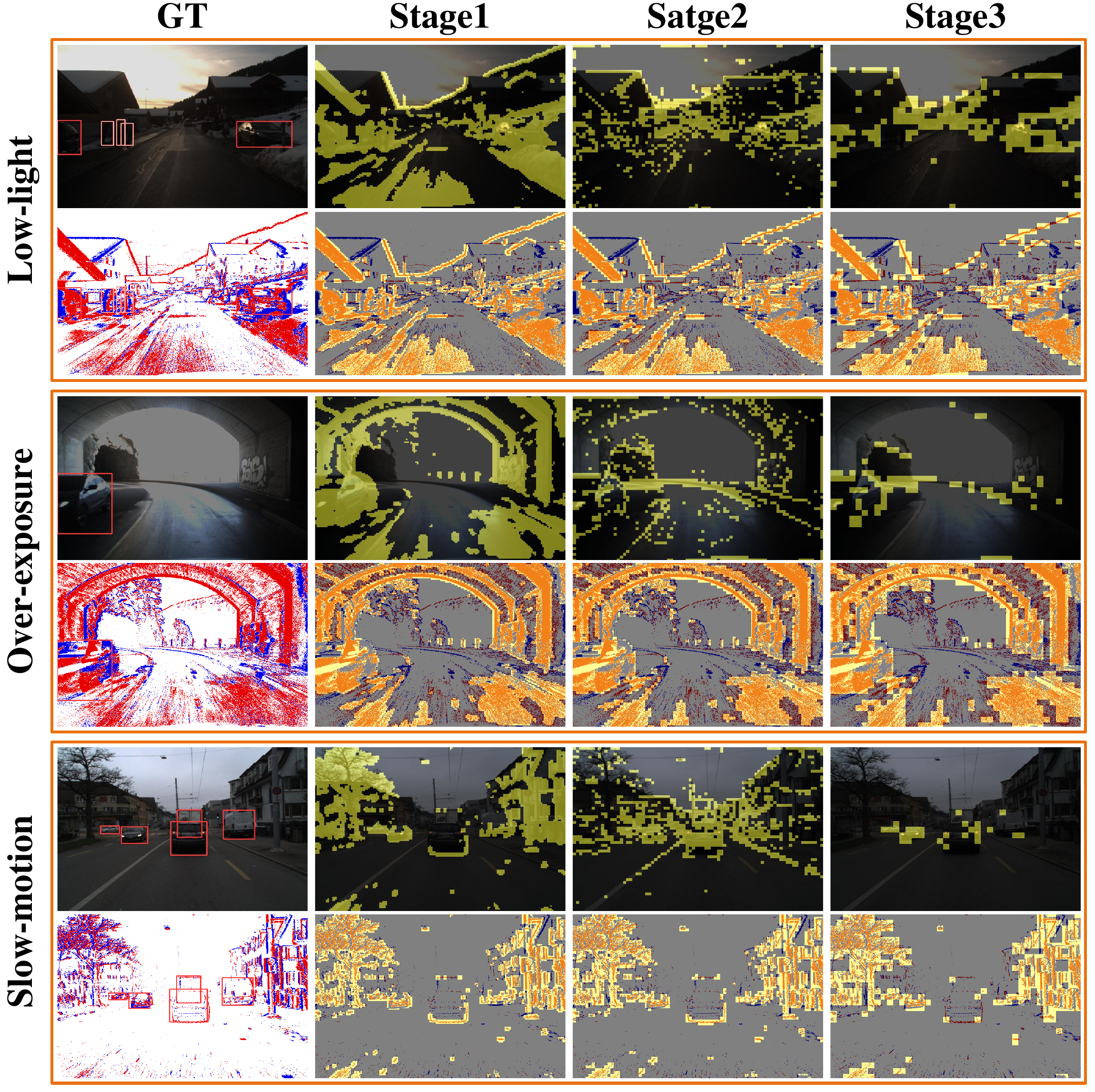}
\caption{Visualizations of sparsification results at different stages on the DSEC-Det dataset.}
\label{fig6}
\end{figure}

\textbf{FI-Mamba module.} We conduct a comparative analysis of various regional focus strategies in the FI-Mamba module. As shown in Tab.\ref{tab:ablate_FIMamba}, IFI-Mamba which focuses on the intersecting regions of the sparsification maps from both modalities, loses critical information and results in the poorest performance. Inter-Mamba which focuses on all regions, introduces substantial background interference and leads to suboptimal performance. In contrast, our method effectively focuses on the important regions while avoiding background interference, achieving the best performance.

\textbf{Extension to the ViT-based framework.} To validate the broader applicability of EGMS, we integrate it with the ViT-based backbone MaxViT \cite{tu2022maxvit} and use simple addition to fuse RGB and event features. As shown in Tab.\ref{vit}, incorporating EGMS into MaxViT achieves 0.5\% gain in mAP and 45.5\% reduction in FLOPs on the DSEC-Det dataset, demonstrating the excellent generalization capability of EGMS.

\begin{figure}[t]
\centering
\includegraphics[scale=0.22]{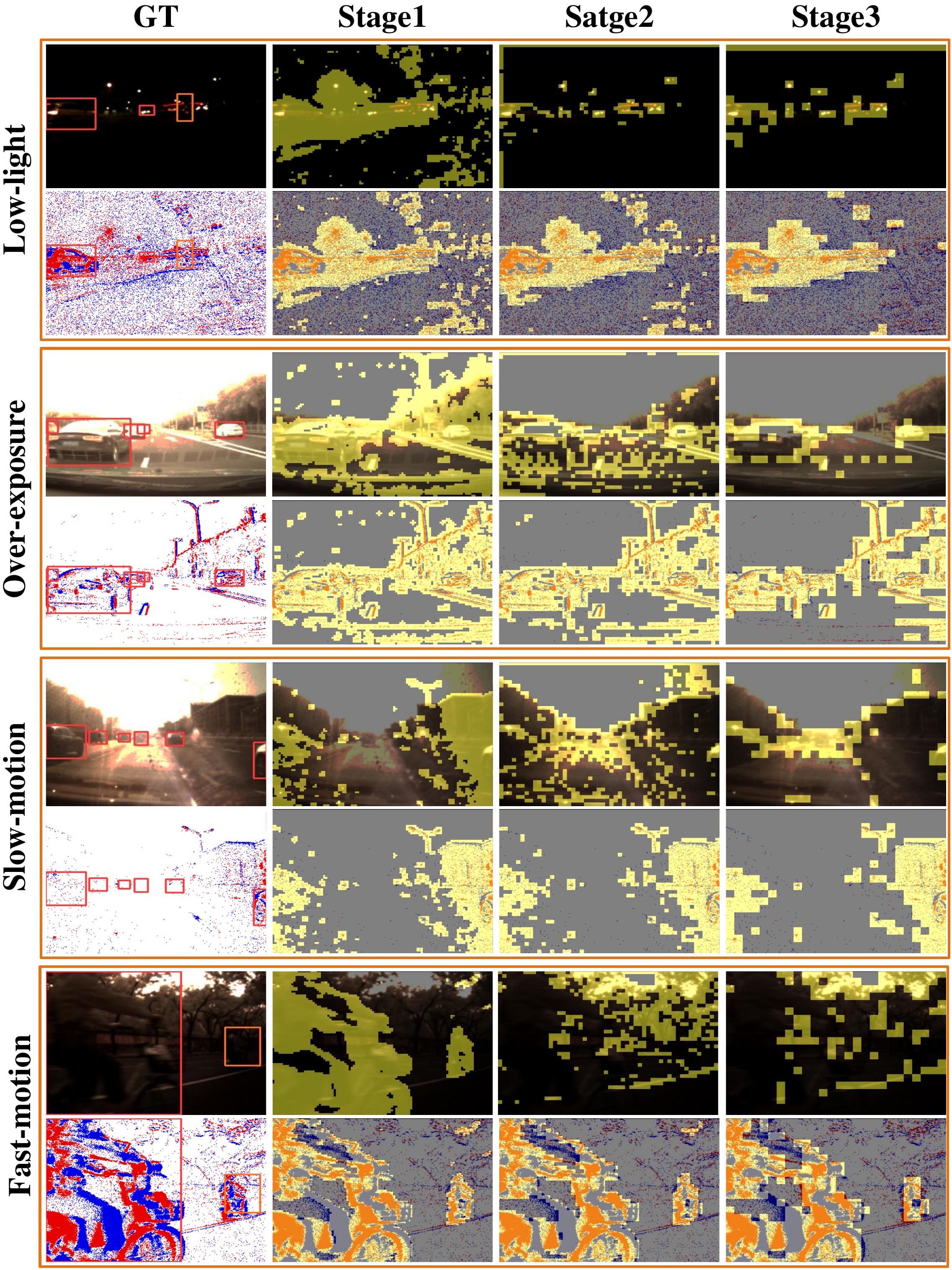}
\caption{Visualizations of sparsification results at different stages on the PKU-DAVIS-SOD dataset.}
\label{fig12}
\end{figure}

\subsection{Sparsification Visualizations}
% The sparsification results for both modalities at different stages on the DSEC dataset are presented in Fig. 6. Under diverse exposure conditions and scene complexities, our EGMS strategy effectively retains crucial information in both modalities, while efficiently eliminating background regions in the image and non-event areas in the event data, thereby demonstrating strong scene-adaptive capabilities.
The sparsification results of the two modalities at different stages on the DSEC-Det and PKU-DAVIS-SOD datasets are shown in Fig.\ref{fig6} and Fig.\ref{fig12}. Under diverse exposure conditions and scene complexities, our EGMS strategy effectively retains informative tokens (yellow regions) for both modalities, while discarding low-quality tokens, demonstrating robust scene-adaptive capabilities. To be noted that the model primarily focuses on image edges in the early layers, resulting in high feature response magnitudes in edge regions, which leads to a failure in identifying important image tokens. Therefore, to preserve critical information in images, the image modality uses the sparsification map from event modality in stage 1.

\begin{figure}[tbb]
\centering
\includegraphics[scale=0.45]{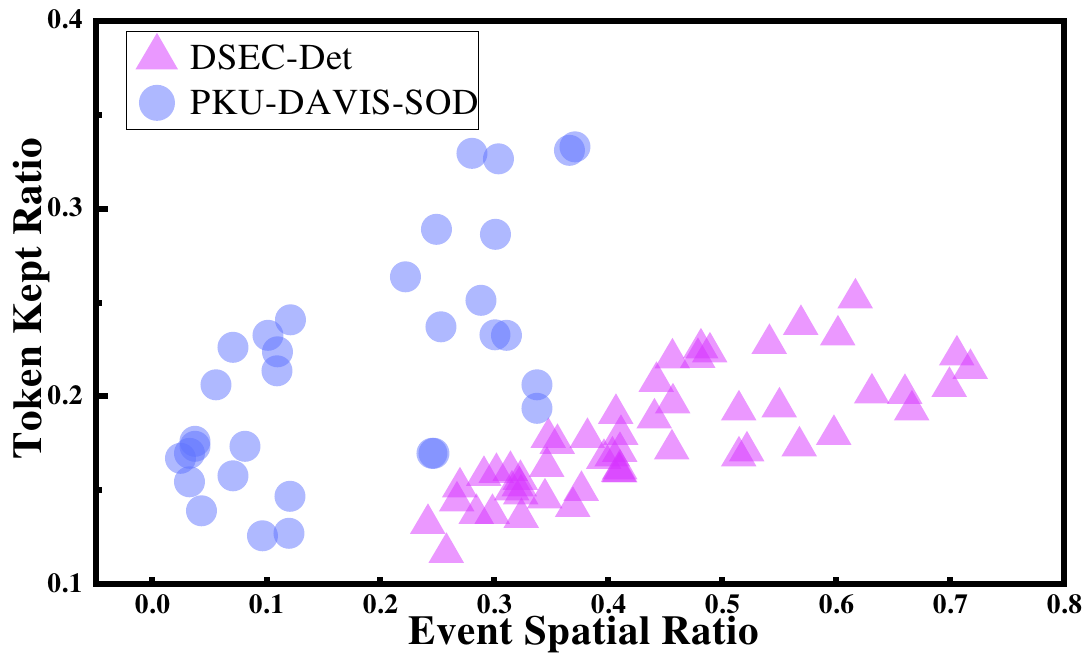}
\caption{The trend of the token kept ratio varies with scene sparsity across the DSEC-Det and PKU-DAVIS-SOD datasets. The observed correlation further demonstrates the scene-adaptability of our FocusMamba. Each point represents a sequence.}
\label{fig13}
\end{figure}

Additionally, we assess the scene sparsity by calculating the ratio of triggered event pixels. The relationship between the token kept ratio and the event spatial ratio across the DSEC-Det and PKU-DAVIS-SOD datasets is depicted in Figure \ref{fig13}. The overall trend depicted in the figure indicates that the token kept ratio increases with the event spatial ratio, demonstrating that our method can adaptively perform sparsification based on the scene's sparsity.

\section{Conclusion}
In this paper, we propose the FocusMamba which achieves an excellent balance between accuracy and efficiency for RGB-Event based object detection. The EGMS strategy efficiently regulates the token scoring and selection process based on the scene’s sparsity level, enabling adaptive retention of critical information and significantly reducing the computational cost. Guided by the sparsification results, the CMFF module effectively perceives and utilizes complementary features, eliminating background interference during the fusion process. Experimental results on two datasets demonstrate that our method achieves superior performance.

\textbf{Limitation.} When the scene and the object remain stationary relative to the camera, the event spatial ratio $r$ is difficult to estimate the object information. In such cases, the control factor approaches 1, losing its regulatory effect on the token selection process. To address this limitation, we will incorporate temporal information in future work to enable the propagation of static object information across frames.

% 当场景和目标相对于相机静止时，事件空间占比接近于0，不能估计出目标的信息量。在这种情况下，控制因子会接近于1，失去对token选择过程的调控。未来，为了在这种情况下对目标的信息量进行更精确的估计，我们会引入时序维度的信息，在帧间传播静止目标的信息

\bibliographystyle{IEEEtran}
\bibliography{trans}

\end{document}